\pdfoutput=1

\documentclass[11pt]{article}

\usepackage[preprint]{acl}

\usepackage{times}
\usepackage{latexsym}

\usepackage[T1]{fontenc}

\usepackage[utf8]{inputenc}

\usepackage{microtype}

\usepackage{inconsolata}

\usepackage{graphicx}
\usepackage{amsmath}

\usepackage[linguistics]{forest} 
\usepackage{synttree}
\usepackage{subcaption}
\usepackage{tikz-dependency}
\usepackage[cjk]{kotex}
\newcommand{\zh}[1]{\begin{CJK}{UTF8}{gbsn}#1\end{CJK}}
\usepackage{langsci-gb4e} 
\usepackage{arydshln}

\newcommand*{\jp}[1]{{\color{black}#1}}

\title{
Parsing Through Boundaries in Chinese Word Segmentation
}

\author{
Yige Chen$^{1,*}$~~~ Zelong Li$^{2,*}$~~~ Cindy Zhang$^{3,*}$~~~ Changbing Yang$^{3,}$\thanks{Equal contribution. $^{\dagger}$Corresponding authors: Haihua Pan and Jungyeul Park.}~~~ \textbf{Zejiao Zeng}$^{3}$\\
\textbf{Amandisa Cady}$^{3}$~~~ \textbf{Ai Ka Lee}$^{3}$~~~  \textbf{Eunkyul Leah Jo}$^{3}$~~~ \textbf{Haihua Pan}$^{1,\dagger}$~~~ \textbf{Jungyeul Park}$^{3,\dagger}$~~~\\
$^{1}$The Chinese University of Hong Kong, Hong Kong\\
$^{2}$University College London, United Kingdom~~~
$^{3}$The University of British Columbia, Canada \\
\url{https://github.com/XDDz123/Chinese_WB_Dependency_Visualization}
\\}

\begin{document}
\maketitle

\begin{abstract}
Chinese word segmentation is a foundational task in natural language processing (NLP), with far-reaching effects on syntactic analysis. Unlike alphabetic languages like English, Chinese lacks explicit word boundaries, making segmentation both necessary and inherently ambiguous. This study highlights the intricate relationship between word segmentation and syntactic parsing, providing a clearer understanding of how different segmentation strategies shape dependency structures in Chinese. Focusing on the Chinese GSD treebank, we analyze multiple word boundary schemes, each reflecting distinct linguistic and computational assumptions, and examine how they influence the resulting syntactic structures. To support detailed comparison, we introduce an interactive web-based visualization tool that displays parsing outcomes across segmentation methods. 
\end{abstract}

\section{Introduction}
Word segmentation plays a crucial role in Chinese natural language processing (NLP), directly impacting downstream tasks such as dependency parsing \citep{wong2022introduction}. Unlike English and other Indo-European languages, which rely on explicit spaces to separate words, Chinese text is written in a continuous sequence of characters. This lack of orthographic word boundaries necessitates the use of segmentation algorithms, each of which imposes different structural interpretations on the linguistic data. 

A central challenge in defining words in Chinese\footnote{Unless otherwise specified, the term \textit{Chinese} in this paper refers to Modern Standard Mandarin Chinese.} is the distinction between words and morphemes. A {word} is typically defined as the smallest grammatical unit that can function independently, whereas a {morpheme} represents the minimal meaningful unit. Many Chinese words are monomorphemic, such as \zh{人} \textit{rén} (`person'), while others are multimorphemic, such as \zh{电脑} \textit{diànnǎo} (`computer'). The challenge of segmentation arises when {free} morphemes, which may function as independent words in some contexts, form compound words in others, such as \zh{出版} \textit{chūbǎn} (`to publish'), composed of \zh{出} \textit{chū} (`to go out; to come out') and \zh{版} \textit{bǎn} (`a block of printing').

Syntactically, words play a fundamental role in structuring dependency relationships. The choice of segmentation scheme directly affects the syntactic analysis of a sentence by altering the number and type of dependency relations. For instance, a {word} such as \zh{向上} \textit{xiàngshàng} (`upward') could be treated as a single adverbial {unit} or split into its component {morphemes} \zh{向} \textit{xiàng} (`toward') and \zh{上} \textit{shàng} (`up'), leading to different dependency tree structures. Similarly, {nominal} compounds such as \zh{中山南路} \textit{Zhōngshān Nánlù} ('Zhongshan South Road') may either be segmented as a single proper noun or decomposed into separate {free} morphemes, which changes the dependency links within the parse tree.


{In this study, we investigate how different segmentation strategies influence dependency structures in Chinese by leveraging the Chinese GSD treebank.\footnote{\url{https://universaldependencies.org/treebanks/zh_gsdsimp/index.html}}} Given the aforementioned challenges, our study evaluates multiple word boundary (WB) segmentation schemes, each providing a distinct perspective on tokenization in Chinese NLP. These segmentation strategies vary in their underlying linguistic assumptions and computational methodologies, impacting downstream tasks such as dependency parsing.

Additionally, we develop a visualization web system to display Chinese parsing results across different segmentation methods. This system will allow users to compare dependency structures generated under different segmentation schemes, highlighting the syntactic differences caused by varying word boundaries. By providing an interactive tool, we aim to support linguistic analysis, facilitate NLP model debugging, and enhance educational applications for Chinese syntactic processing.

\section{Morphology of Chinese}


The Chinese language originated as a monomorphemic language. In Old Chinese, almost all words were monosyllabic and monomorphemic, meaning that each word consisted of only one morpheme and one character \citep{dong2020history}. While Chinese morphology has undergone drastic changes over the past few centuries, evolving from Classical Chinese to Modern Chinese, the characteristics of its morphemes have remained largely unchanged: most morphemes in Chinese continue to be monosyllabic and consist of a single character \citep{myers2022wordhood}.

On the other hand, word formation in Chinese differs significantly from that in Classical Chinese, and multimorphemic words are now common in the language. While Chinese, as an isolating language, has significantly less morphology than agglutinating and polysynthetic languages \citep{liao2014morphology}, it allows certain derivational and inflectional affixes to attach to stems to form multimorphemic words. Unlike many other languages, such as English, Chinese does not mark tense or parts of speech morphologically \citep{sun2006chinese}.

\subsection{Affixes in Chinese}

\paragraph{Derivational morphemes}
Nominal, verbal, and adjectival derivational affixes occur in Chinese \citep{liao2014morphology}. A typical nominal morpheme is the suffix \zh{学} \textit{xué} (`study'), which commonly denotes an academic discipline when attached to a stem. For example, the word \zh{化学} \textit{huàxué} (`chemistry') consists of \zh{化} \textit{huà} (`to change into') and \zh{学} \textit{xué} (`study'), conveying the meaning `study of change'.

\paragraph{Inflectional morphemes}
The Chinese language utilizes inflectional morphemes to mark aspects. In Mandarin Chinese,\footnote{The use of aspectual suffixes can differ in other Chinese variants such as Cantonese and Shanghainese.} the aspectual suffixes include the perfective \zh{了} \textit{le}, experiential \zh{过} \textit{guò}, and progressive \zh{着} \textit{zhe}. A verbal morpheme (typically a free morpheme that can stand alone as a word), such as \zh{学} \textit{xué} (`to study'), may take these aspectual suffixes to indicate the perfective (\zh{学了} \textit{xuéle}), experiential (\zh{学过} \textit{xuéguò}), or progressive (\zh{学着} \textit{xuézhe}) aspect. Other inflectional morphemes, such as the plural suffix for humans (\zh{们} \textit{men}), also exist in Chinese.

\subsection{Chinese compounds}

Another prominent type of word formation in Chinese is compounding. Since Chinese generally lacks derivational and inflectional morphemes compared to synthetic languages, compounding is a major source of Chinese words. Compounds consist of two or more (free or bound) roots \citep{chao1965grammar}, which distinguish them from derivational or inflectional morphemes attached to a root. As a result, the formation of Chinese compounds is largely driven by the semantics of the stems \citep{song2022compounding}. The parts of speech of the two formants of Chinese compounds vary, including verbs, nouns, adjectives, adverbs, etc. Chinese compounds can be classified into the following types \citep{liao2014morphology}: 

\paragraph{Coordinative/parallel compounds} In this compound type, the two constituent morphemes or roots carry similar, related, or complementary meanings. Both elements typically belong to the same grammatical category. An example is \zh{帮助} \textit{bāngzhù} (`to help', verb), which combines \zh{帮} \textit{bāng} (`to help') and \zh{助} \textit{zhù} (`to assist'), both verbal formants with closely aligned meanings.

\paragraph{Modifier-head compounds}
These compounds involve an internal relationship in which the first morpheme functions as a modifier that constrains or qualifies the meaning of the second morpheme, which serves as the head. For instance, \zh{慢跑} \textit{mànpǎo} (`to jog', verb) is composed of \zh{慢} \textit{màn} (`slow', adjective) modifying \zh{跑} \textit{pǎo} (`to run', verb), forming a compound denoting a type of slow running.

\paragraph{Verb-resultative compounds}
This type consists of two morphemes where the first expresses an action and the second denotes the result or outcome of that action. For example, \zh{晒干} \textit{shàigān} (`to sun-dry', verb) is made up of \zh{晒} \textit{shài} (`to sun', verb) and \zh{干} \textit{gān} (`dry', adjective), where the drying is the intended result of the sunning process.

\paragraph{Subject-predicate compounds}
In these compounds, the first morpheme acts as the subject, and the second serves as the predicate. For example, \zh{脸红} \textit{liǎnhóng} (`to blush', verb) combines \zh{脸} \textit{liǎn} (`face', noun) with \zh{红} \textit{hóng} (`red', adjective), denoting the face becoming red.

\paragraph{Verb-object compounds} One root refers to the predicate, while the other root is a thematically related object. For instance, \zh{出版} \textit{chūbǎn} (`to publish', verb) consists of \zh{出} \textit{chū} (`to go out; to come out', verb) and \zh{版} \textit{bǎn} (`a block of printing', noun).


\paragraph{Issues in Chinese compounds}
Given that a Chinese compound consists of two formants, it is typically the case that one functions as the head of the compound, determining its syntactic category. It has been reported that nearly 90\% of compound nouns have a nominal formant on the right, and 85\% of compound verbs have a verbal formant on the left \citep{packard2000morphology}. Consequently, for a disyllabic compound, the default pattern is for the head to appear on the left if the compound is verbal, and on the right if it is nominal \citep{sun2006chinese}.

Compounds in Chinese also exhibit the syntax of the language. For compounds consisting of a verbal formant and a nominal formant serving as the object of the verb (V-O compounds), the positions of the two formants resemble the syntactic positions of a verb and its object in the predicate-argument structure. While the Lexical Integrity Hypothesis assumes that syntactic transformations are not applicable to word-internal structures, \citet{huang1984phrase} observes that Chinese V-O compounds can permit syntactic processes to affect internal parts of the V-O sequence. Moreover, for compounds consisting of two verbal formants (V-V compounds), \citet{li1990vv} observes that they can be accounted for by standard Case theory alongside assumptions of Government-Binding Theory.

Another interesting observation regarding Chinese compounds is the phenomenon of separable words (\zh{离合词} \textit{líhé-cí}). It seems that in some cases, the two morphemes in a V-O compound can be separated, allowing words or phrases to appear between them. For example, the compound \zh{帮忙} \textit{bāngmáng} (`to help; to do a favor') consists of \zh{帮} \textit{bāng} (`to help') and \zh{忙} \textit{máng} (`busy'). The two morphemes can be split to form phrases like \zh{帮他的忙} \textit{bāng-tāde-máng} (`to do him a favor'), where \zh{他的} \textit{tāde} (`his') is inserted. It is plausible to assume that syntax is involved in this process, in a way that the verbal morpheme is free and functions like a verb, and the nominal morpheme is also free and functions like a noun. In fact, studies such as \citet{huang2008his} adopt this view and favor an analysis in which the two formants occupy separate syntactic positions. On the other hand, this clearly violates the Lexical Integrity Hypothesis. An alternative analysis proposed by \citet{pan2015liheci} considers the so-called `separation' an epiphenomenon. The seemingly separated morphemes are cognate object constructions, where a verb and a noun of the same underlying form occupy the two positions but are reduced to their heads in surface form. The phrase \zh{帮他的忙} \textit{bāng-tāde-máng} (`to do him a favor') is therefore analyzed as \zh{帮忙他的帮忙} \textit{bāngmáng-tāde-bāngmáng} underlyingly, with the nominal formant \textit{máng} in the verb and the verbal formant \textit{bāng} in the noun being deleted, reducing to its surface representation. This analysis prevents syntax from intervening in word-internal structures and keeps the Lexical Integrity Hypothesis intact. 

Overall, the morphology of Chinese sheds light on word segmentation in several aspects: (1) Inflectional morphemes may be considered separate tokens from stems, as their stems are always free; (2) Given the possible involvement of syntax in Chinese compounds, it may or may not be legitimate to assert that some compounds (especially V-O compounds) can be split into distinct tokens. While it remains debatable whether syntax plays a role in word formation, differences in word segmentation undoubtedly affect syntactic analysis, given the ambiguities at the boundary between morphology and syntax in Chinese.

\section{Segmentation Strategies for Chinese}

\subsection{Word segmentation variants}

Given the central role of word segmentation as a preprocessing step in Chinese NLP and the complexity of defining word boundaries in Chinese, we investigate how different word boundary systems impact downstream syntactic structures.
We consider the following segmentation strategies, each offering a distinct perspective on how Chinese texts can be tokenized:

\paragraph{Morpheme-based segmentation}
The Chinese GSD treebank mainly employs a morpheme-based segmentation approach.\footnote{We use the term \textit{morpheme} in a functional sense to refer to the minimal meaning-bearing units identified in the Chinese GSD treebank, acknowledging that some segmentations may not correspond to traditional linguistic morphemes.}
This method focuses on breaking text down into minimal meaningful units, making it particularly useful for fine-grained linguistic analysis. Morpheme-based segmentation is beneficial for tasks requiring precise syntactic and morphological distinctions, though it may sometimes introduce challenges in recovering full-word meaning.

\paragraph{Word-based segmentation} Language Technology Platform (LTP) provides a word-based segmentation model that aligns more closely with traditional lexicon-driven approaches \citep{che-etal-2010-ltp,che-etal-2021-n}. This method treats compound words and idiomatic expressions as unified entities, improving syntactic coherence in downstream parsing tasks. LTP's approach ensures a more natural representation of words as they function in linguistic contexts, making it well-suited for structured NLP applications such as syntactic parsing and machine translation.

\paragraph{Corpus-based segmentation}
The Penn Chinese Treebank (CTB) \citep{xue-EtAl:2005} and the Peking University Modern Chinese Corpus (PKU) \citep{shiwen-etal-2002-pekin} segmentation, implemented in CoreNLP \citep{manning-etal-2014-stanford}, follows a data-driven approach that balances linguistic motivation with corpus-based learning. This segmentation strategy aims to reflect linguistic intuition by incorporating manually annotated corpora, producing a segmentation scheme that aligns closely with human expectations. 

\subsection{Empirical comparison}

We compare three segmentation schemes, namely LTP, CTB, and PKU, against the original GSD segmentation to examine how different definitions of word boundaries affect Chinese text processing. While all segmenters aim to tokenize raw Chinese sentences, each reflects distinct assumptions about lexical units, morphological transparency, and the treatment of compound expressions and named entities.

\paragraph{GSD versus LTP}

The LTP segmentation favors lexicalized expressions and holistic treatment of multimorphemic compounds. For instance, expressions like \zh{天文台} \textit{tiānwéntái} (`astronomical observatory') are treated as single words in LTP, whereas GSD segments them into \zh{天文} \textit{tiānwén} (`astronomy') and \zh{台} \textit{tái} (`platform'), reflecting a modifier-head structure. Named entities such as \zh{医学人文博物馆} \textit{yīxué rénwén bówùguǎn} (`Medical Humanities Museum') appear as single tokens in LTP but are segmented morpheme-by-morpheme in GSD, as \zh{医学} \textit{yīxué} (`medical'), \zh{人文} \textit{rénwén} (`humanities'), \zh{博物} \textit{bówù} (`museum-related'), and \zh{馆} \textit{guǎn} (`hall'). Temporal and locative phrases follow a similar pattern: GSD segments \zh{2004年} \textit{èr líng líng sì nián} (`year 2004') as \zh{2004} and \zh{年} \textit{nián}, and \zh{中山南路} \textit{Zhōngshān Nánlù} (`Zhongshan South Road') as \zh{中山} \textit{Zhōngshān}, \zh{南} \textit{nán} (`south'), and \zh{路} \textit{lù} (`road'), while LTP treats these expressions as single, lexicalized units. These differences show how LTP prioritizes surface-level lexical cohesion, whereas GSD emphasizes morphological decomposability.

\paragraph{GSD versus CTB}

The CTB segmentation exhibits similar tendencies to LTP, especially in its treatment of nominal compounds and named locations. Like LTP, CTB frequently treats expressions such as \zh{中山南路} \textit{Zhōngshān Nánlù} and \zh{天文社} \textit{tiānwénshè} (`astronomy club') as single tokens, while GSD segments them into \zh{中山} \textit{Zhōngshān}, \zh{南} \textit{nán}, \zh{路} \textit{lù}, and \zh{天文} \textit{tiānwén}, \zh{社} \textit{shè} (`club'). Institutional names such as \zh{医学人文博物馆} \textit{yīxué rénwén bówùguǎn} are also lexicalized in CTB but decomposed in GSD. CTB consistently reflects preferences for readability and surface coherence, aligning with common usage. However, subtle variations distinguish CTB from LTP. For example, in the treatment of foreign terms or abbreviations, CTB sometimes segments more aggressively than LTP. While both diverge from GSD in similar directions, CTB introduces its own segmentation conventions in specific domains.

\paragraph{CGD versus PKU}

PKU segmentation reflects a highly lexicalized approach, often merging tokens that GSD segments for morphological or syntactic clarity. Numeral-classifier expressions like \zh{2004年} \textit{èr líng líng sì nián} are consistently joined in PKU but segmented in GSD. Terms such as \zh{亚热带} \textit{yàrèdài} (`subtropical') are treated as atomic units in PKU, while GSD separates them into \zh{亚} \textit{yà} (`sub-') and \zh{热带} \textit{rèdài} (`tropical zone'). Similarly, building names like \zh{博物馆} \textit{bówùguǎn} (`museum') are kept intact in PKU but segmented as \zh{博物} \textit{bówù} and \zh{馆} \textit{guǎn} in GSD. These patterns demonstrate PKU's emphasis on established lexical forms and contrast with GSD's preference for structure-preserving tokenization.

\paragraph{Summarization}
Across all three comparisons, a consistent trend emerges: GSD prioritizes linguistic transparency through morphological and syntactic segmentation, while LTP, CTB, and PKU emphasize lexical cohesion. The differences commonly involve nominal compounds, named entities, and affixal structures. These segmentation strategies reflect different definitions of wordhood in Chinese and have important implications for tasks such as parsing and named entity recognition. Although each segmentation scheme has its strengths depending on the application context, the lack of a shared gold standard complicates direct evaluation. 

\section{Implementation and Discussion}

\paragraph{WB conversion process in GSD} \label{Conversion}
The conversion process takes as input the Chinese GSD CoNLL-U files (\texttt{GSDSimp} from Universal Dependencies), along with separate pre-segmented versions of each sentence and their corresponding predicted universal part-of-speech (UPOS) tags. Using a word alignment function grounded in pattern-matching alignment \citep{jo-etal-2024-untold-story}, the system compares the original GSD tokenization with an alternative segmentation that permits the combination of adjacent words. Crucially, this process respects the integrity of the original GSD tokens, as no token is split to form a new unit. To maintain structural consistency, combinations that would introduce multiple dependency-head relations across token boundaries are disallowed, unless they form legitimate words or reflect word boundaries not marked in the original GSD segmentation. The transformation is implemented directly on the GSD tokens by concatenating relevant fields such as lemma and language-specific part-of-speech (XPOS) tags, in accordance with the CoNLL-U format. UPOS labels for the newly combined tokens are drawn from the predicted annotations, which follow an alternative tagging scheme. Whitespace is removed between combined tokens, except in cases involving foreign words. Once the process is complete, the data structures are re-indexed and sorted for export back into CoNLL-U format.

\begin{figure*}[!th]
\centering
\resizebox{.9\textwidth}{!}{
\includegraphics{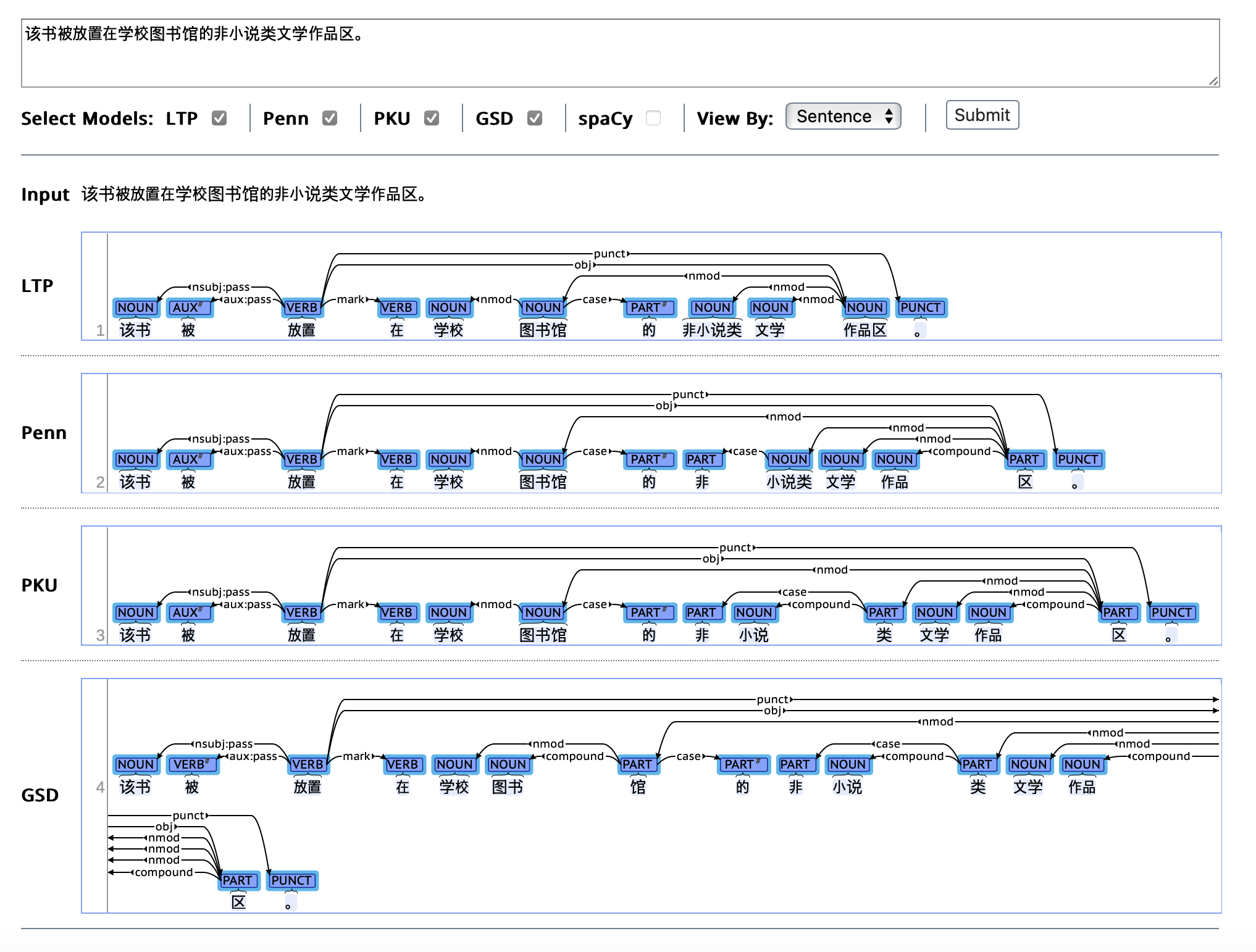}
}
\caption{Screenshot of the visualization system using \texttt{brat} with the example sentence \textit{Gāi shū bèi fàngzhì zài xuéxiào túshūguǎn de fēi xiǎoshuō lèi wénxué zuòpǐn qū.} (`The book was placed in the nonfiction literature section of the school library.'). 
}
    \label{visualization}
\end{figure*}

\paragraph{Systems}
We implement our system using \texttt{stanza} \citep{qi-etal-2020-stanza}. Each dependency parser is trained on a different version of the Chinese GSD treebank, where each version has been pre-segmented using a distinct word segmentation strategy. This allows us to isolate and compare the syntactic effects introduced by segmentation variation within a consistent parsing framework. We visualize the resulting dependency structures using \texttt{brat} \citep{stenetorp-etal-2012-brat}. Parsed sentences are converted into the \texttt{brat} annotation format, enabling side-by-side inspection of structural differences across segmentation schemes.
Figure~\ref{visualization} illustrates how a sentence
can yield different dependency structures depending on the word segmentation scheme.


\paragraph{Experimental results}
We evaluate both dependency parsing and word segmentation performance across different segmentation schemes. 
Results are reported using standard Universal Dependencies metrics, including Unlabeled Attachment Score (UAS) and Labeled Attachment Score (LAS), allowing us to assess how segmentation differences impact syntactic structure.
However, the number of tokens varies across segmentation schemes due to differing word boundary definitions. As a result, direct comparison of parsing scores may not be entirely fair, since each scheme imposes a different structural granularity on the data. A rigorous evaluation would require aligning the outputs to a shared representation or normalizing performance metrics across schemes. We leave such a fair and systematic comparison for future work.\footnote{One potential approach to enable fair comparison is to normalize all outputs by splitting compound words according to the original GSD segmentation. This would ensure that the number of tokens remains consistent across different schemes, allowing for more direct comparison of dependency structures. However, such an approach introduces its own limitations: the alternative segmentation outputs often treat compound words as single units, and splitting them post hoc would discard internal dependency relations that were never explicitly annotated. As a result, comparing parsing accuracy on such normalized representations could misrepresent the syntactic intent of the original segmentation. Future work may explore alignment-based evaluation methods or develop shared annotation frameworks that abstract away from token-level granularity while preserving structural comparability, as seen in constituency parsing evaluation \citep{park-etal-2024-jp}.}

For word segmentation, we evaluate both the predictions generated by \texttt{stanza} and the outputs from the original segmentation tools used in the creation of the datasets. Since no gold-standard segmentations are available across all schemes, we rely entirely on predicted outputs—either reproduced by our implementation or generated by the original tools themselves. This approach allows us to assess not only the internal consistency of each segmentation method, but also the reproducibility of each scheme when applied in practice. Our goal is to evaluate how reliably each model reflects its intended segmentation strategy when reapplied to the same data.
Table~\ref{results} presents the dependency parsing results (UAS/LAS) and word segmentation reproducibility (F$_1$ scores) across the different segmentation schemes.

\begin{table}[!ht]
\centering
\footnotesize{
\begin{tabular}{c|c c c c } \hline 
         &  GSD & LTP & CTB & PKU \\ \hline 
UAS         & 80.05 & 76.13 & 74.51 & 73.21\\
LAS         & 77.11 & 72.88 & 71.38 & 70.17\\ \hdashline
WB          & 93.98 & 92.51 & 89.03 & 86.89 \\
\hline 
    \end{tabular}}
    \caption{Dependency parsing results (UAS/LAS) and word segmentation reproducibility (F$_1$).}
    \label{results}
\end{table}

\paragraph{Discussion}

Our experimental results highlight the significant impact that word segmentation has on dependency parsing performance in Chinese. Variations in segmentation schemes lead to differences in the number and structure of tokens, which in turn affect both the syntactic representations learned by the parser and the evaluation metrics used to assess parsing quality. While \texttt{stanza} demonstrates stable parsing performance across most segmentation schemes, small but consistent fluctuations in UAS and LAS suggest that certain segmentation strategies may better align with the syntactic assumptions of the UD annotation framework. The word segmentation F$_1$ scores further reflect the challenge of reproducing original segmentation schemes using predictive models. Although \texttt{stanza} performs competitively, discrepancies between segmentation tools, such as LTP, CTB, and PKU, indicate differences in underlying linguistic assumptions about wordhood. These discrepancies underscore the importance of segmentation reproducibility when evaluating downstream tasks like dependency parsing. In general, our findings confirm that segmentation is not a neutral preprocessing step in Chinese NLP but a linguistically consequential decision that shapes syntactic analysis. Effective parsing of Chinese text depends not only on the parsing model itself but also on the consistency and linguistic validity of the segmentation scheme used in training and evaluation.

\section{Conclusion}
This study presents a comparative analysis of dependency structures produced by different word segmentation schemes, offering insights into how segmentation strategies influence syntactic representations in Chinese NLP. \jp{Broader implications of Chinese word segmentation can also be explored in future work.\footnote{For example, see Appendix~\ref{gec-wb}.}} Additionally, we develop a visualization tool to support linguistic analysis and model debugging.

\section*{Limitation}
This study focuses on Chinese dependency parsing using the GSD treebank and a limited set of word segmentation tools. Our assumption is that GSD reflects the finest segmentation granularity among the available resources, and we construct coarser granularities by merging its tokens accordingly. While this allows for a controlled comparison across segmentation strategies, the assumption may not hold for other treebanks in Universal Dependencies, which often adopt different segmentation guidelines. As such, our findings are constrained by the scope of models and datasets used. 
Future work could extend this analysis to other treebanks, explore additional languages with similar challenges, and investigate segmentation-aware parsing models.

\section*{Ethics and Broader Impact Statement}
This work uses publicly available datasets and tools, including the Chinese GSD treebank from Universal Dependencies and open-source NLP frameworks such as \texttt{stanza}. No private or sensitive data were used. The analysis focuses solely on linguistic structure and does not involve human subjects or personally identifiable information. All experiments were conducted in accordance with academic standards for reproducibility and transparency.


\appendix

\section{Licenses}
Our datasets and trained models, derived from the Universal Dependencies (UD) Chinese GSDSimp Treebank, are publicly available and distributed under the Creative Commons Attribution-ShareAlike 4.0 International License (CC BY-SA 4.0). This license allows users to share (copy and redistribute the material in any medium or format) and adapt (remix, transform, and build upon the material for any purpose, including commercial use), provided that proper attribution is given to the original dataset and that any derivative works are released under the same license.

\section{GEC Annotation under Different Word Boundary Schemes} \label{gec-wb}

Grammatical error correction is a downstream task that is highly sensitive to word segmentation choices, particularly in Chinese, where word boundaries are not overtly marked. Our extended analysis illustrates how different segmentation schemes shape not only tokenization but also the annotation of learner language.

Word boundaries play a crucial role in Chinese language processing and education, yet their ambiguity presents challenges for both natural language processing systems and language learners. This paper investigates the impact of WB segmentation on grammatical error annotation in L2 Chinese writing, aiming to identify the most effective WB strategies for supporting learners and instructors. We examine various segmentation approaches, comparing outputs produced by different NLP models to assess their influence on error detection and classification.

We refine the Chinese \texttt{errant} framework by incorporating a new error typology that accounts for phonetic, visual, and structural similarities in character misuses. Additionally, we adapt word-based segmentation schemes to improve annotation accuracy, mitigating tokenization inconsistencies found in the Chinese GSD treebank. A perception experiment involving L2 learners and instructors further evaluates how WB variations affect grammatical error identification and pedagogical effectiveness.

Our primary implementation builds on the \texttt{jp-errant} framework for Chinese GEC \citep{wang-etal-2025-refined}, which performs edit-based alignment and error categorization using part-of-speech tags and token-level comparisons. By default, \texttt{jp-errant} employs \texttt{stanza} \citep{qi-etal-2020-stanza} with GSD-style segmentation, which adopts a fine-grained, morpheme-oriented notion of wordhood. This approach captures subtle syntactic and morphological errors but can also fragment multiword expressions, complicating certain types of error interpretation.

To enable a more direct comparison with prior work such as \texttt{ChERRANT} \citep{zhang-etal-2022-mucgec}, which is based on the LTP analyzer \citep{che-etal-2010-ltp}, we provide a \texttt{stanza}-based reimplementation of LTP-style segmentation. In contrast to GSD, the LTP scheme prioritizes lexical cohesion, grouping compounds and fixed expressions into single tokens. This often aligns more closely with learner intuitions and yields clearer boundaries for identifying lexical and syntactic errors in educational settings.

These findings suggest that word segmentation is not merely a preprocessing step but a structural choice that significantly influences error interpretation and system evaluation. The broader implications of segmentation granularity extend to annotation protocols, evaluation metrics, and the design of learner-aware NLP tools for Chinese.

Having both segmentation variants allows us to isolate the role of word boundaries in shaping annotation outcomes. For example, under GSD-style segmentation, a compound like \zh{中山南路} \textit{Zhōngshān Nánlù} (`Zhongshan South Road’) may generate multiple edit operations if only one character is erroneous. In contrast, under LTP-style segmentation, the entire compound may be treated as a single token, yielding a different classification. These structural differences affect both alignment and labeling, ultimately influencing how system performance is assessed.
By supporting both GSD-style and LTP-style segmentation in our GEC annotation pipeline, we promote more robust and interpretable evaluations of GEC systems for Chinese. 


\end{document}